\documentclass[runningheads]{llncs}
\usepackage{amsmath}
\usepackage{graphicx}
\usepackage{textcomp}
\usepackage[dvipsnames]{xcolor}
\usepackage{amssymb}
\usepackage{comment}
\usepackage{multirow}
\usepackage{adjustbox}
\usepackage{authblk}
\newcommand{\etal}{\textit{et al.}}

\begin{document}
\title{Diffusion-based Document Layout Generation}
%
%
\author{Liu He\inst{1}\thanks{Work done while a research intern at Microsoft Cloud and AI.} \and
Yijuan Lu\inst{2}\and
John Corring\inst{2}\and
Dinei Florencio\inst{2}\and
Cha Zhang\inst{2}}
\authorrunning{He et al.}
%
\institute{Purdue University, West Lafayette IN 47906, USA\\
\email{he425@purdue.edu}\\
\and Microsoft Cloud and AI, Bellevue WA 98004, USA\\
\email{\{yijlu,jocorrin,dinei,chazhang\}@microsoft.com}}
\maketitle              

\begin{abstract}
We develop a diffusion-based approach for various document layout sequence generation. Layout sequences specify the contents of a document design in an explicit format. Our novel diffusion-based approach works in the sequence domain rather than the image domain in order to permit more complex and realistic layouts. We also introduce a new metric, Document Earth Mover's Distance (\textit{Doc-EMD}). By considering similarity between heterogeneous categories document designs, we handle the shortcomings of prior document metrics that only evaluate the same category of layouts. Our empirical analysis shows that our diffusion-based approach is comparable to or outperforming other previous methods for layout generation across various document datasets. Moreover, our metric is capable of differentiating documents better than previous metrics for specific cases.
\end{abstract}
\section{Introduction}
Document creation involves many steps from generating textual content, organizing additional media, and producing a layout that makes the information comprehensible. Layout generation is a key step in document creation. Layouts differ to best convey the appropriate kind of information for different domains of documents. To model many different domains of document layouts, general yet powerful methods need to be developed. We undertake that goal in this paper.  

Methodological and domain considerations for layout generation have arisen as a topic of interest recently in the Computer Vision and Machine Learning communities~\cite{layoutgan,patil2020read,gupta2021layouttransformer}. Fixed length generative methods leveraging adversarial training were shown effective at producing realistic but limited documents in LayoutGAN~\cite{layoutgan}. In READ~\cite{patil2020read} an approach to resolve shortcomings of the prior work was developed, namely permitting more complex structures in the layout, as well as introducing document metric considerations in the generative model literature. This work proposed a recursive autoencoder to iteratively extend the layout based on past generated layouts. However, READ~\cite{patil2020read} relies on a hierarchical document model that may not apply well in a wide range of document types, and is dependent on hyperparameters that dictate document layout length.  This work can be seen as a segway to autoregressive methods such as Layout Transformer~\cite{gupta2021layouttransformer} where the modern generative techniques of language modeling have been applied to the field. Autoregressive generation is outlined in more detail below. This method allows for adaptive stopping that comes from the encoder states themselves, which change progressively during the generation. The result is simpler and improved modeling of layouts.

In this paper, we develop a new approach to layout generation using the recently emerging area of diffusion probabilistic models. The key idea is that when a diffusion process consists of small steps of Gaussian noise conditioned on the data, then the reversing process can be approximated by a conditional Gaussian as well. To use the conventional diffusion methods for discrete sequence generation, rounding and embedding steps have to be introduced~\cite{DiffusionLM2022}. 

Our main contributions are introducing a novel document comparison metric with several useful properties and being the first work to employ discrete-sequence diffusion for layout generation. We show that the proposed metric behaves well qualitatively and quantitatively by comparing the performance of different algorithms on well-known datasets. We also show how synthetic layout training data compares to real document data on an end-to-end task: layout detection. We compare across different synthetic data generation algorithms by the mean average precision (mAP) of a trained layout detector of a fixed architecture. Finally, we provide ablation of the proposed method to several variables.

\section{Related Works}

\subsection{Generative Networks}
Generative adversarial networks~\cite{goodfellow2020generative} launched the generative revolution in image generation~\cite{denton2015deep,isola2017image,radford2015unsupervised,karras2019style}, and text generation~\cite{yu2017seqgan,che2017maximum,guo2018long}. This self-supervised training scheme enables the networks to consume large unlabeled realistic dataset, and provides a powerful baseline in various downstream tasks like image colorization~\cite{nazeri2018image}, image compositing~\cite{zhu2021barbershop}, and text synthesis~\cite{li2016precomputed}. Variational autoencoder~\cite{kingma2013auto} is a counterpart framework in generative network domain. The network excludes the burden of discriminator and using only variational loss and regularization loss on latent space in the bottleneck. The concise of both network and training scheme support its generalized representation of the real dataset ~\cite{van2017neural,he2022masked}. 

Modern generative methods have a root in autoencoding pretext tasks from the NLP literature~\cite{devlin2018bert}. Masked language modeling is a category of pretraining tasks in which pieces of a token sequence are hidden with a \verb|[MASK]| token and the model fills in the missing piece. In the autoregressive setting, this approach is simplified to a forward-looking token regression and so models generally have a backward-looking attention emphasis, generating the newest token sequentially. This autoregressive scheme achieves success in computer vision tasks like ViT~\cite{dosovitskiy2020image}, large language models~\cite{radford2018improving} like GPT-3~\cite{brown2020language}.

\subsection{Layout Generation}
Numerous works focusing layout generation have been proposed recently. LayoutGAN~\cite{li2020layoutgan} and LayoutVAE~\cite{jyothi2019layoutvae} provide general 2D layout planning for natural images. More works focus on document layout generation~\cite{kikuchi2021constrained,tabata2019automatic,lin2020variational}. In particular, the work of ~\cite{patil2020read} introduced recursive autoencoding for layout generation, as well as a layout similarity \textit{DocSim}, which looks at the geometric similarity of layouts weighted by size of the elements.~\cite{gupta2021layouttransformer} is an example, in which the transformer architecture is applied in this autoregressive fashion to generate layout tokens (class labels and bounding boxes). In~\cite{arroyo2021variational} ideas from the transformer-based methods mentioned above are combined with variational autoencoders to produce a more controllable and predictable generator. This work also implements the Wasserstein sequence distance as a metric into the layout literature. Our work builds on this idea and gives the Wasserstein distance more geometric significance, giving us the \textit{Doc-EMD} metric that compares well with the Wasserstein and the \textit{DocSim}.

\subsection{Diffusion Generative Methods}
Diffusion models use a sequential denoising model as an objective to generate realistic objects from Gaussian noise~\cite{ho2020denoising,NicholDiffusion2021,SongDenoising2020}. A domain-specific mean-estimator is used to model a markov process that is similar in spirit to deconvolution, but can be trained in an end-to-end manner. This approach is primarily applied in the image domain due to the approximately continuous nature of images and the denoising approach relying on using a Gaussian estimator for the inversion step. The diffusion model has achieved great success in text-image synthesis~\cite{ramesh2022hierarchical,rombach2022high,gu2022vector}, 3D neural rendering~\cite{poole2022dreamfusion}, 3D point cloud generation~\cite{luo2021diffusion}, image compositing~\cite{song2022objectstitch}, audio generation~\cite{kong2020diffwave,agostinelli2023musiclm}, and video generation~\cite{molad2023dreamix}. Recent works have shown that to extend this generative mechanism to discrete spaces, a rounding network can be applied that maps the denoised token embedding to a dictionary~\cite{DiffusionLM2022}. Our work builds on this and applies this approach to document layout, producing realistic document layouts that can be conditioned on partial layouts.

\section{Methods}

\begin{figure}
\centering
\includegraphics[width=0.8\linewidth]{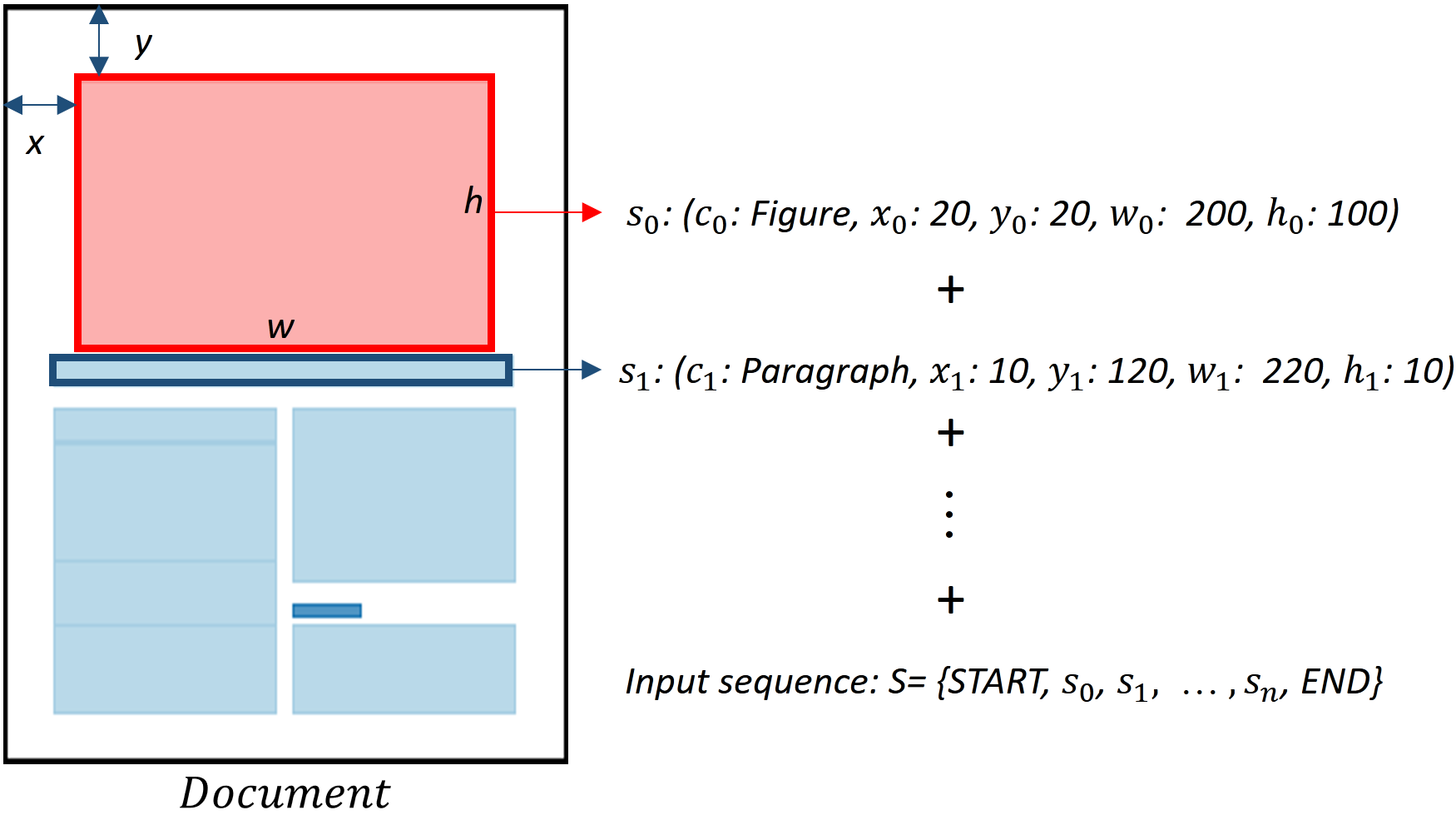}
\caption{ \bf{Sequence representation of document layout.} } \label{fig: seq rep}
\end{figure}

\subsection{Diffusion for Layout Generation}

Our approach leverages recent advances in denoising diffusion applied to discrete sequences ~\cite{SongDenoising2020,DiffusionLM2022}. As Figure~\ref{fig: seq rep} shows, we consider a document layout of $N $ layout elements given by the $5-$tuples $S = \{s_i\}_{i=1}^N = \{(c_i, x_i, y_i, w_i, h_i)\}_{i=1}^N$, where $x_i$ is the upper-left $x-$coordinate, $y_i$ is the upper-left $y-$coordinate, $w_i, h_i$ specify the width and height of the box (all in pixel coordinates), and $c_i$ specifies the class (table, figure, formula, etc.) of the $i-$th box. Note that $x_i,y_i,w_i,h_i \in \mathbb{N}$ and $c \in [K]$ where $K$ is the number of layout classes. To process this data structure as a sequence, we serialize $S$ by flattening the sequence $\{c_0, x_0, y_0, w_0, h_0, c_1, \ldots \}$. Then the geometric entries are discretized and quantized to a fixed vocabulary $G$ of size $|G|$. We can offset the class index entries by $|G|$ to keep the class coordinates distinct, or simply use a unique token outside of the vocabulary for $G$. This yields a sequential representation of length $5N$ with a fixed vocabulary $V = G \cup [K]$. 

\begin{figure}
\includegraphics[width=\linewidth]{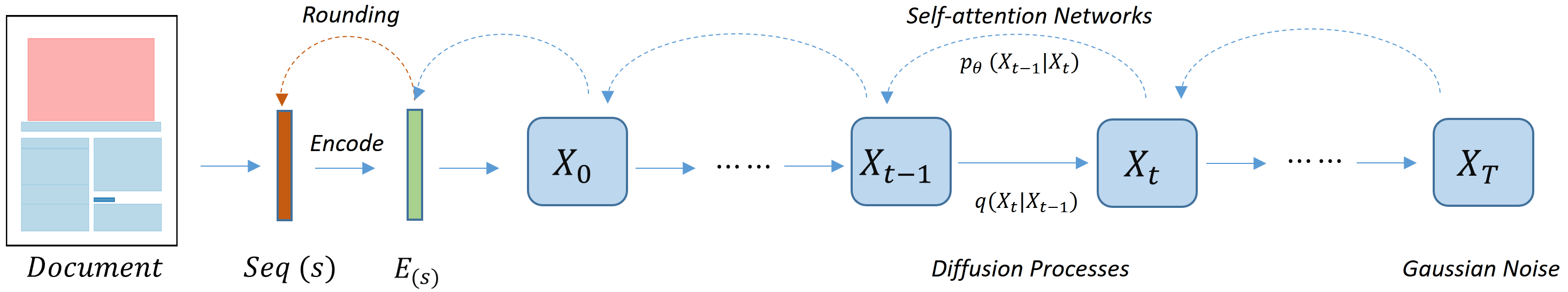}
\caption{ \bf{Diffusion for Layout Generation.} }
\label{fig: diffsion model}
\end{figure}

The framework of our method is illustrated in Fig.~\ref{fig: diffsion model}. An embedding step $E$ is introduced to map the discrete sequence $s \in V^N$ to a feature vector $E(s) \in \mathbb{R}^{d \times N}.$ We extend the conventional denoising transition model $x_t \rightarrow x_{t-1} \rightarrow \ldots x_0$ with another transition $x_0 \rightarrow E(s)$. Recall that the learned parameters refine the transition probability estimates:
\begin{align}
p_\theta (x_{t-1} \vert x_t) = \mathcal{N}(x_{t-1} ; \mu_\theta (x_t; t), \Sigma_\theta (x_t, t))
\end{align}
by minimizing the variational upper bound of the negative log likelihood of the image over $\theta$. The analogy to sequential learning is the log likelihood of the sequence embedding, so we take $x_0 \sim E(s)$ following~\cite{DiffusionLM2022}. To close the loop with the original sequence, a rounding module is introduced $p_\theta(s | x_0)$ which estimates the token sequence from the embedding $E(s) \approx x_0.$ This is done by a rounding function with learned biases, call it $R$. So the (bidirectional) Markov chain is: 
\begin{align}
&s \overset{E}{\rightarrow} w \overset{q}{\rightarrow} x_1 \overset{q}{\rightarrow} \ldots x_T, \notag \\
&s \overset{R}{\leftarrow} w \overset{p_\theta}{\leftarrow} x_1  \overset{p_\theta}{\leftarrow} \ldots x_T.
\end{align}

To solve for the reverse-process parameters several key simplifications were devised by ~\cite{ho2020denoising}. The end result was to reduce the full variational bound objective to a simpler mean squared loss. The initial variational bound objective is written
\begin{equation}
   L = \min_\theta \mathbb{E}  \left( -\log (p_\theta(x_T)) - \sum_{t \geq 1} \log\frac{p_\theta (x_{t-1} | x_t)}{q(x_t | x_{t-1})} \right).
\end{equation}
This can be rewritten as a series of KL-Divergences ~\cite{ho2020denoising}
\begin{equation}
   \min_\theta \mathbb{E}  \left( D_{KL}(q(x_T |x_0) || p_\theta(x_T)) + \sum_{t>1} D_{KL}(q(x_{t-1} | x_t, x_0) || p_\theta (x_{t-1} | x_t)) \right) \notag \\
     - \log (p_\theta (x_0 | x_1) ) 
\end{equation}
and each term is a divergence between two Gaussian distributions except for the last. By regrouping terms and throwing out the constant $\log( x_T )$ term, the closed form divergence for each term results in a mean squared error loss
\begin{equation}
    L_2 = \sum_{t=1}^{T} \mathbb{E} \| \mu_{\theta} (x_t, t) - \hat{\mu} (x_t, x_0) \|^2. 
\end{equation}
$\hat{\mu}$ is the mean of the posterior and $\mu_\theta$ is the predicted mean of the reverse process computed by a model parameterized by $\theta$. In our implementation we use a transformer as $\mu_\theta$ and $T=2000$. Note that for $\hat{\mu}$ a closed form gaussian, the derivation of which can be found in~\cite{ho2020denoising}.

\subsection{\textit{Doc-EMD}: Earth Mover's Distance as a Document Metric} \label{intro docemd}

Document metrics are nontrivial because of the complex nature of documents~\cite{patil2020read}. Many metrics and similarities have been proposed, all with variation shortcomings that defy intuitive reasoning about the nature of document layouts. We propose to leverage the Earth-Movers distance, which has been deployed successfully in contour matching~\cite{grauman2004fast} as well as image matching~\cite{rubner2000earth}, to provide an underlying distance for document layouts. This allows us to leverage some useful properties of this well-established distance and we can leverage high quality existing open-source implementations to implement the Earth-Mover's distance~\cite{flamary2021pot}. 

\subsubsection{Definitions and Formal Distance}

Consider a layout of $N $ layout elements given by the $5-$tuples $S = \{s_i\}_{i=1}^N$. We call this the source layout, and $T= \{t_i\}_{i=1}^{M}$ is the target layout. In this metric, we consider layouts as consisting of the $2-$d points in the integer pixel coordinates falling inside of each layout box. Consider a single layout box $s=(c,x,y,w,h)$ we take $s \cong \rho(S) = \{ p : x < p_1 < x+w, y < p_2 < y+h\}$ which we take as equivalent to the uniform pointwise density generated by those points. Let $|\cdot|$ denote the number of point elements in this set. Then we define the earth-mover's distance as 
\begin{align}
EMD(s,t) &= EMD(\rho(s), \rho(T)) = \min_{F} \sum_{i,j} F_{ij} d(s_i, t_j) \\
& s.t. \sum_{j} F_{i,j} \leq 1/{|s|}, \sum_{i} F_{i,j} \leq 1/{|t|}, \sum_{i,j} F_{i,j} = 1. \notag 
\end{align}
We conflate the EMD between class bounding box representation of $s$ with $\rho(s)$ so that the cost of matching the element scales up with the size of the element. 
So now we have defined the data structure of the overarching layout $S$, each layout element $s$, and an element-to-element distance $EMD$ which allows us to compare elements from different layouts. 

Now we define the distance between two layouts $S$ and $T$. First define the class function $C(s = (c,x,y,w,h)) = c$ and the set function $\hat{C}(S, c) = \{s \in S : C(s) = c\}$. Let $\kappa(cls; S, T)$ be the indicator function as to whether only one of $S$ and $T$ has elements of class $cls$ (the exclusive or). The \textit{Doc-EMD} is defined as 
\begin{align}
DOC_\lambda(S,T) &=\sum_{cls}  EMD(\cup_{\hat{C}(S,cls)} s, \cup_{\hat{C}(T,cls)} t) + \sum_{cls : \kappa(cls; S, T) = 1} \lambda,
\end{align}
where $\lambda$ is some positive factor used to penalize missing classes (we use $\lambda=1$).
In language, $DOC$ is the sum of the earth-mover's distances between the sets of elements in $S$ and $T$ belonging to the same class plus a penalty term for each class that only appears in $S$ or $T$. Note that this is very different from the \textit{DocSim}~\cite{patil2020read} since we can avoid the computation of the size weightings in favor of the pointwise pmf contributions as well as skip the Hungarian matching step in favor of the earth-movers matching. Meanwhile, our method adds substance to the layout semantics which is missing from the Wasserstein sequence metrics~\cite{arroyo2021variational} as they only capture exact matchings in the location and weight location and class in a disproportionate manner.

$DOC$ is clearly reflexive, symmetric, and positive. Note that each term consisting of the $EMD$ on subsets of $S, T$ is a metric. Then for the second term, note that this is 
 the discrete metric on the projection of the pointset to the class it belongs to scaled by the penalty term $\lambda$. So $DOC_\lambda$ is a sum of metrics, which is also a metric, so it obeys the triangle inequality. This proof sketch shows that it enjoys all of the formal advantages of the Wasserstein distance.

\subsubsection{Qualitative Comparisons with other Document Metrics and Similarities}

In this work we use several metrics to evaluate the performance of our proposed algorithm. We begin with a comment that document similarity (or distance) is a nontrivial problem. Documents are complex objects that can be represented in a number of ways and typically have no canonical underlying space from which they are drawn. Modeling that is the goal of generative layout algorithms, but how to measure the quality is directly related to the complexity of this problem. There is no perfect solution, so we develop an approach that covers some of the shortcomings of existing metrics, which we discuss now.

First, \textit{DocSim}~\cite{patil2020read} proposes a Hungarian-matching based algorithm that uses a weighting term that scales linearly in the minimum area between boxes and exponentially in the size and shape difference. It has several major shortcomings
\begin{enumerate}
    \item Not having an open-sourced common implementation;
    \item Not well normalized so may not compare well between datasets;
    \item The "similarity" is not an inner-product, so may not behave well in some cases.
\end{enumerate}
We show qualitative examples highlighting each of these shortcomings.

\begin{figure}
\includegraphics[width=\linewidth]{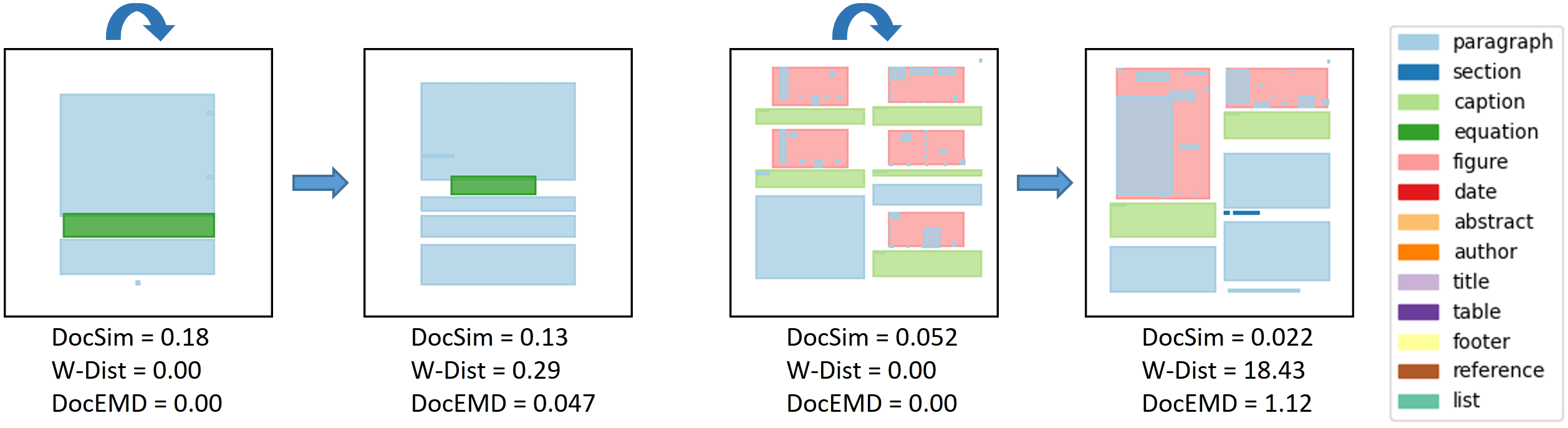}
\caption{ \bf{DocSim Similarities.} Visualization of how the \textit{DocSim}, Wasserstein Distance, and \textit{Doc-EMD} Distance vary across pairs of documents. The left image represents the source and the subsequent images are the targets. The metrics applied between them are shown below the target. The metrics shown below the source are the self-comparisons. \textit{DocSim} is not normalized, has no common "self-similarity" value (this depends on the document), and does not always vary intuitively across structures. Wasserstein Distance is overly sensitive to class mismatch and does not capture structural similarities well. Note that this metric is primarily meant for distribution comparison. Also, note that as it is a distance higher values mean farther away (unlike the similarity of \textit{DocSim}). For \textit{Doc-EMD} Distance, there is a clear separation between single- and multi-column when classes are similar, the distance is well bounded, and the self-comparison is standardized.}
\end{figure}

Second, the Wasserstein sequence distance is applied to the layout sequences as opposed to the layout geometry. This acts as a distributional distance between the empirical discrete class label (categorical) distribution and the continuous bounding box distribution (Gaussian). First, this does not explicitly model similarity between two documents well (here it suffers from confusion across variations of box instances in the individual layouts), but rather between the distributions of documents given from a corpus. This is the key strength of this distance: it is a natural measure of distributional match. This is why we see this used commonly in generative modeling ~\cite{arjovsky2017wasserstein}. Note that it does not allow for weighting by area in any conventional sense, an advantage of the \textit{DocSim} which is lost here.


Finally, the \textit{Doc-EMD} distance is applicable both at the document pair level and at the distribution level. To compute the distributional similarity we take a similar approach to \textit{DocSim}. We first compute the pairwise distance matrix with our metric applied to each pair of images. Then we obtain a negative matching score to which we apply the Hungarian matching algorithm. Key advantages over the former approaches are: 
\begin{enumerate}
    \item It is well normalized, scaling from 0 to K (the class number) if the appropriate pixel coordinates are used;
    \item It keeps the distance properties of the Wasserstein without losing the geometric specificity of \textit{DocSim};
    \item It behaves predictably when comparing structurally and semantically different layouts due to the geometric specificity of the per-class EMD.
\end{enumerate}
Perhaps the greatest weakness of our method is that it requires greater runtime than the previous two in the pair-wise distance stage. However, as mentioned above, using open-sourced packages with hardware acceleration as well as optional speedups from approximations make it feasible even for very large datasets ~\cite{flamary2021pot}.

\section{Experiments}\label{sec:experiments}
\subsection{Dataset}
We evaluate our method on various public available document datasets from journal articles, tables, to magazines.

\textbf{PubLayNet}~\cite{zhong2019publaynet} consists of 330K document layouts by matching XML representations of public PDF articles from PubMed Centra\textsuperscript{TM}. It has 5 semantic categories including \textit{Text, Title, List, Figure,} and \textit{Table}. Semantic elements on each layout are annotated by categorized bounding box in COCO~\cite{lin2014microsoft} format. Typically there is no overlapping between semantic units. We utilized its official splits: 335,703 for training, 11,245 for validation.

\textbf{DocBank}~\cite{li2020docbank} consists of 500K document layouts by weak supervision of articles available on the arXiv.com. It contains 12 categories including \textit{Abstract, Author, Caption, Equation, Figure, Footer, List, Paragraph, Reference, Section, Table,} and \textit{Title}. Its annotated bounding boxes are created by merging extraction results of text lines. In general, this dataset has more (number) and fragmented (size) annotated bounding boxes compared to PubLayNet~\cite{zhong2019publaynet}. And overlapping exists between semantic units (\textit{Paragraph} within \textit{Figure} etc.). We use 399,811 layouts for training, and 49,980 layouts for validation.


\textbf{Magazine}~\cite{zheng2019content} has 4K magazine layouts classified into 6 semantic categories including \textit{Text, Image, Headline, Text-over-image, Headline-over-image,} and \textit{Background.} The dataset itself holds natural overlaps between categories.

\subsection{Comparisons}
We compare our approach with 3 related methods on document layout generation task, including LayoutVAE~\cite{jyothi2019layoutvae},  Gupta~\etal~\cite{gupta2021layouttransformer}, and VTN~\cite{arroyo2021variational}. We utilize the code provided by the author's repository of~\cite{gupta2021layouttransformer} for the first two approaches. In particular, we privilege the ground truth bounding box count to LayoutVAE and only train its BBoxVAE portion. For Gupta~\etal~\cite{gupta2021layouttransformer}, the original inference code use the first bounding box as input prior and sample the top-$k$ ($k=5$) predicted layout bounding boxes to enable diversity. We modify the inference code to generate from initial token rather than the first bounding box. During inference, we find that results' diversity vanishes fast with $k<5$. Thus the top-5 sampling is kept to ensure fairness. For VTN~\cite{arroyo2021variational}, we add the variational training scheme to the code of \cite{gupta2021layouttransformer}. For fairness, all methods are trained for the same epochs on each dataset. Their models are based on default settings in the code or recommended settings in the original papers.

\subsubsection{Quantitative Results} \label{sec: quantitative}
For all three datasets we list, we generate 1000 ( 391 for Magazine dataset since its validation dataset is small. ) document layouts by ours and other competitors. Then we compare the generated results with the same amount of real document layouts by series of metrics. Specifically, the LayoutVAE~\cite{jyothi2019layoutvae} is conditioned on an input bounding box count. Gupta~\etal~\cite{gupta2021layouttransformer} and VTN~\cite{arroyo2021variational} generation is fully random from initial token. 

Our set-by-set comparison is evaluated by 4 quantitative metrics. We already discuss the capability of \textit{Doc-EMD} in Section~\ref{intro docemd}, and we also include \textit{DocSim} as reference. In theory, these two metrics should indicate reverse pattern (Smaller \textit{DocSim} corresponds to larger \textit{Doc-EMD}. And our results illustrate this pattern.). \textit{Overlap} is the percentage of total overlapping area among the generated layout bounding boxes. Generally, less overlap indicates better performance a method achieves. However, we notice that there is a certain amount of reasonable overlapping existing in DocBank ( Legend texts in a figure are recognized as paragraph. Thus its bounding box overlaps with figure's bounding box, etc. ) and Magazine dataset ( The categories of \textit{Text-over-image}, and \textit{Headline-over-image} naturally overlap with \textit{Image} category. ). Thus reasonable amount ($<2\%$, etc.) of overlapping area will not affect the realism of generated layouts in terms of DocBank and Magazine dataset. Finally, \textit{Coverage} is the percentage of total bounding box area over the document extent area. The closer value to the real data one, the better performance a method achieves.

As shown in Table~\ref{tabpubley}, ~\ref{tabdocbank}, ~\ref{tabmagazine}, our method outperforms competitors in most metrics on 3 datasets. Specifically, our method achieves the best performance in \textit{DocSim} and \textit{Doc-EMD}, and the second in other two metrics. For DocBank, our method achieves the best performance in \textit{Coverage} and \textit{Doc-EMD}, and the second in \textit{DocSim}. For Magazine, our method dominates the \textit{DocSim}, \textit{Doc-EMD}, and \textit{Overlap}. Moreover, our \textit{Doc-EMD} metric keeps stable scalability and capability across three datasets, which verify the contributions in Section~\ref{intro docemd}.

\begin{table}[t]
\begin{center}
\caption{Benchmark performance on PubLayNet Dataset}\label{tabpubley}
\begin{tabular}{|l|c|c|c|c|}
\hline
Approaches &  \textit{DocSim} $\uparrow$ & \textit{Doc-EMD} $\downarrow$  & Overlap$\downarrow$ & Coverage\\
\hline
Layoutvae~\cite{jyothi2019layoutvae} & 0.129 & 0.191 &  2.02\% & \textbf{56.21\%}\\
 Gupta~\etal~\cite{gupta2021layouttransformer} & 0.137 & \underline{0.063} & \textbf{0.065\%} & 51.62\%\\
 VTN~\cite{arroyo2021variational} & \underline{0.141} & 0.068 & 0.083\% & 53.49\% \\
 Ours & \textbf{0.163} & \textbf{0.053}  & \underline{0.062\%} & \underline{55.30\%} \\
\hline
 Real Data & & & 0.026\% & 56.09\% \\
\hline
\end{tabular}
\end{center}
\end{table}

\begin{table}[!htbp]
\begin{center}
\caption{Benchmark performance on DocBank Dataset}\label{tabdocbank}
\begin{tabular}{|l|c|c|c|c|}
\hline
Approaches &  \textit{DocSim}$\uparrow$ & \textit{Doc-EMD}$\downarrow$ &  Overlap$\downarrow$ & Coverage\\
\hline
Layoutvae~\cite{jyothi2019layoutvae} & 0.087 & 0.592 & 2.02\% & 56.21\%\\
 Gupta~\etal~\cite{gupta2021layouttransformer} & 0.078 & 0.518 &  \textbf{0.56\%} & 44.01\%\\
 VTN~\cite{arroyo2021variational} & \textbf{0.096} & \underline{0.353} & \underline{0.61\%} & \underline{44.27\%} \\
 Ours & \underline{0.093} & \textbf{0.319} & 2.04\% & \textbf{45.49\%} \\
\hline
 Real Data & & & 0.45\% & 46.20\% \\
\hline
\end{tabular}
\end{center}
\end{table}

\begin{table}[!htbp]
\begin{center}
\caption{Benchmark performance on Magazine Dataset}\label{tabmagazine}
\begin{tabular}{|l|c|c|c|c|}
\hline
Approaches &  \textit{DocSim}$\uparrow$ & \textit{Doc-EMD}$\downarrow$ & Overlap & Coverage\\
\hline
Layoutvae~\cite{jyothi2019layoutvae} & \underline{0.260} & 0.143 &  \underline{2.23\%} & \underline{80.93\%}\\
 Gupta~\etal~\cite{gupta2021layouttransformer} & 0.176 & 0.227 &  12.6\% & 81.64\%\\
 VTN~\cite{arroyo2021variational} & 0.232 & \underline{0.138} &  5.29\% & \textbf{79.88\%} \\
 Ours & \textbf{0.302} & \textbf{0.117} &  \textbf{1.23\%} & 70.55\% \\
\hline
 Real Data & & & 1.36\% & 76.00\% \\
\hline
\end{tabular}
\end{center}
\end{table}


\subsubsection{Qualitative Results}
\begin{figure*}[t]
\centering
\setlength{\tabcolsep}{3pt}
\tiny
\begin{tabular}{p{0.08mm}cccccc}
    \rotatebox[origin=l]{90}{\textbf{LayoutVAE~\cite{jyothi2019layoutvae}}}  &
    \includegraphics[width=0.15\textwidth]{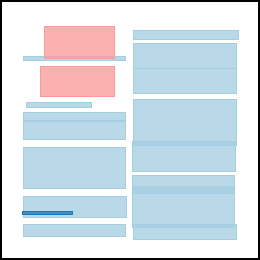} &
    \includegraphics[width=0.15\textwidth]{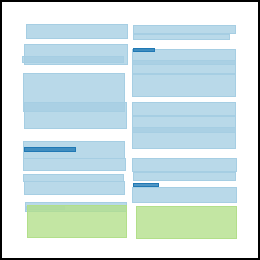} &
    \includegraphics[width=0.15\textwidth]{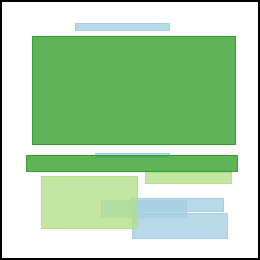} &
    \includegraphics[width=0.15\textwidth]{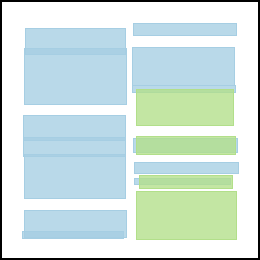} &
    \includegraphics[width=0.15\textwidth]{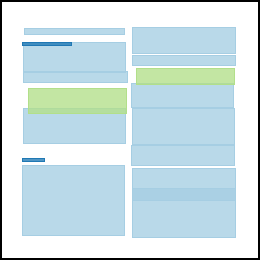} &\\

    \rotatebox[origin=l]{90}{\textbf{Gupta~\etal~\cite{gupta2021layouttransformer}}}  &
    \includegraphics[width=0.15\textwidth]{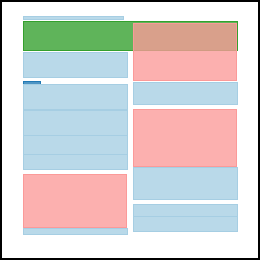} &
    \includegraphics[width=0.15\textwidth]{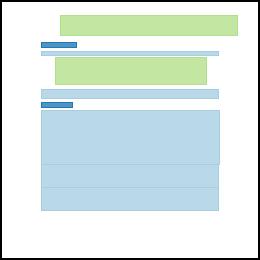} &
    \includegraphics[width=0.15\textwidth]{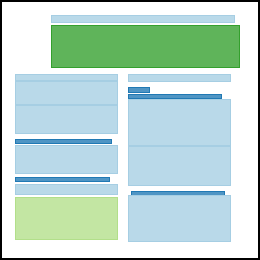} &
    \includegraphics[width=0.15\textwidth]{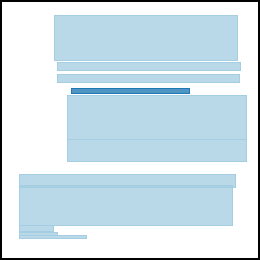} &
    \includegraphics[width=0.15\textwidth]{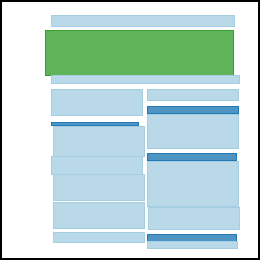} &\\

    \rotatebox[origin=l]{90}{\textbf{VTN~\cite{arroyo2021variational}}}  &
    \includegraphics[width=0.15\textwidth]{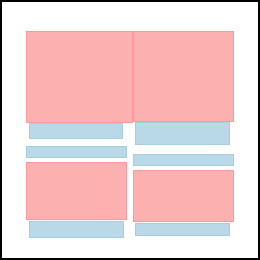} &
    \includegraphics[width=0.15\textwidth]{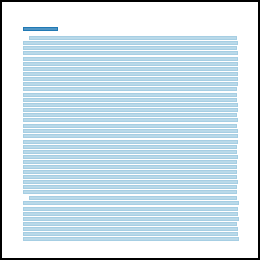} &
    \includegraphics[width=0.15\textwidth]{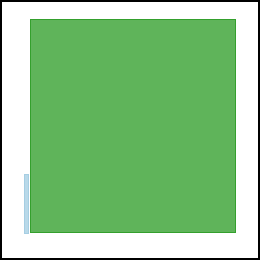} &
    \includegraphics[width=0.15\textwidth]{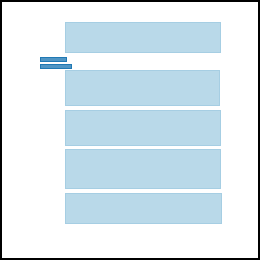} &
    \includegraphics[width=0.15\textwidth]{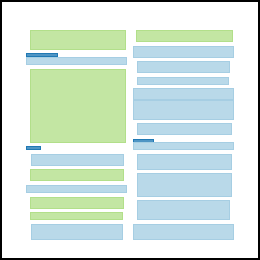} &\\

    \rotatebox[origin=l]{90}{\textbf{Ours}} &
    \includegraphics[width=0.15\textwidth]{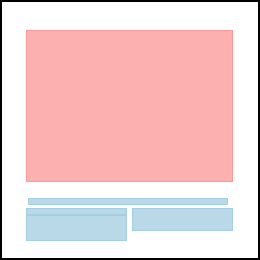} &
    \includegraphics[width=0.15\textwidth]{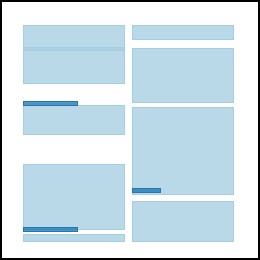} &
    \includegraphics[width=0.15\textwidth]{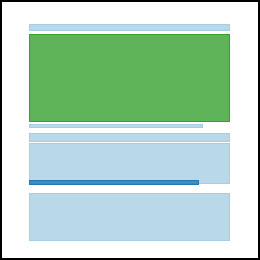} &
    \includegraphics[width=0.15\textwidth]{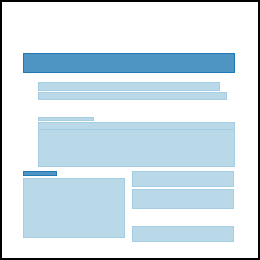} &
    \includegraphics[width=0.15\textwidth]{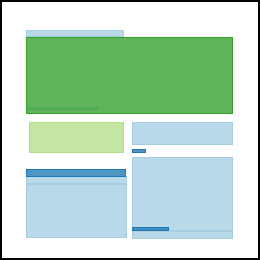} &
    \includegraphics[width=0.08\textwidth]{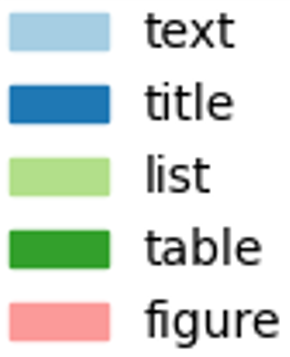} \\
\end{tabular}%
    \caption{\textbf{Qualitative Results for PubLayNet.} Competitors are worse in alignment, repeat similar patterns, contain abnormal bounding box categories, or have noticeable overlaps. Our method appears to represent PubLayNet better.}
  \label{fig:publaynet}
\end{figure*}

\begin{figure*}[!htbp]
\centering
\setlength{\tabcolsep}{3pt}
\tiny
\begin{tabular}{p{0.08mm}cccccc}
    \rotatebox[origin=l]{90}{\textbf{LayoutVAE~\cite{jyothi2019layoutvae}}}  &
    \includegraphics[width=0.15\textwidth]{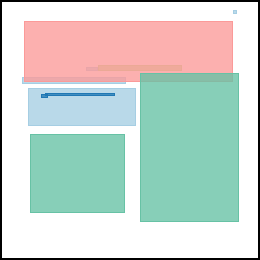} &
    \includegraphics[width=0.15\textwidth]{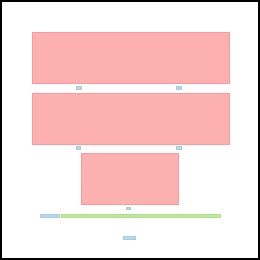} &
    \includegraphics[width=0.15\textwidth]{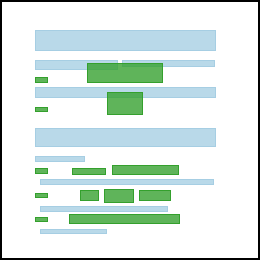} &
    \includegraphics[width=0.15\textwidth]{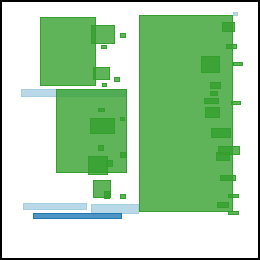} &
    \includegraphics[width=0.15\textwidth]{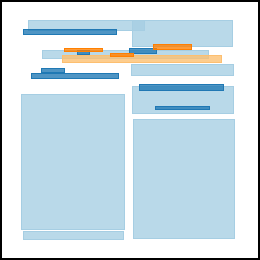} & \\

    \rotatebox[origin=l]{90}{\textbf{Gupta~\etal~\cite{gupta2021layouttransformer}}}  &
    \includegraphics[width=0.15\textwidth]{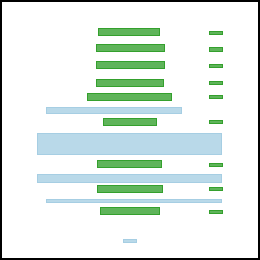} &
    \includegraphics[width=0.15\textwidth]{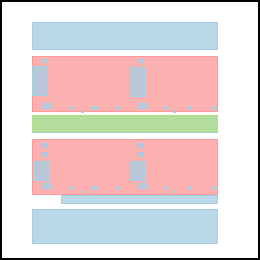} &
    \includegraphics[width=0.15\textwidth]{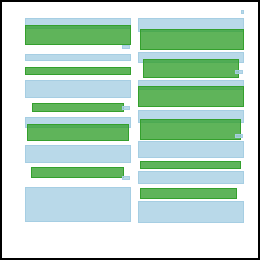} &
    \includegraphics[width=0.15\textwidth]{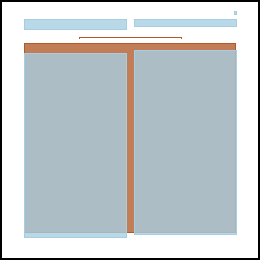} &
    \includegraphics[width=0.15\textwidth]{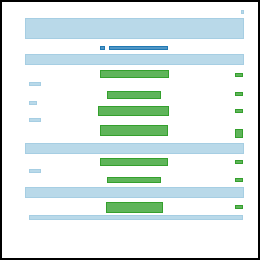} & \\

    \rotatebox[origin=l]{90}{\textbf{VTN~\cite{arroyo2021variational}}}  &
    \includegraphics[width=0.15\textwidth]{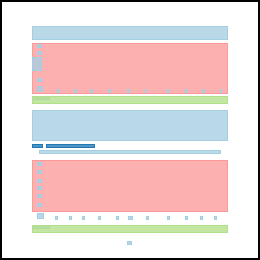} &
    \includegraphics[width=0.15\textwidth]{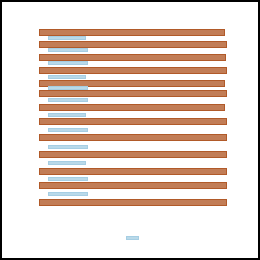} &
    \includegraphics[width=0.15\textwidth]{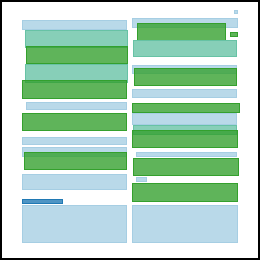} &
    \includegraphics[width=0.15\textwidth]{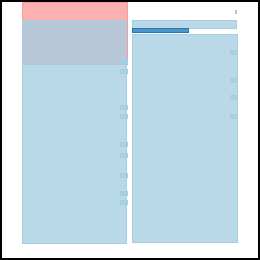} &
    \includegraphics[width=0.15\textwidth]{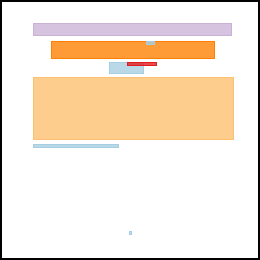} & 
    \multirow{2}{0.09\textwidth}{\includegraphics[width=0.08\textwidth]{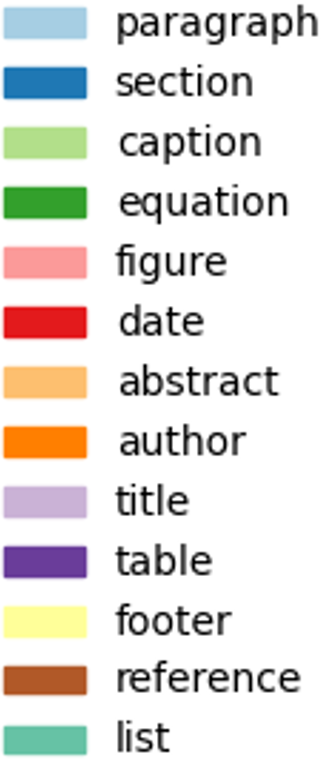}}\\

    \rotatebox[origin=l]{90}{\textbf{Ours}}  &
    \includegraphics[width=0.15\textwidth]{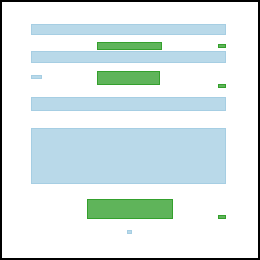} &
    \includegraphics[width=0.15\textwidth]{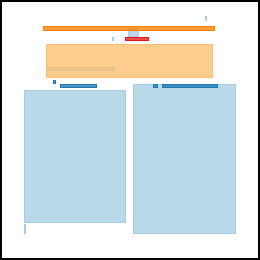} &
    \includegraphics[width=0.15\textwidth]{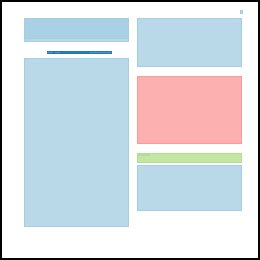} &
    \includegraphics[width=0.15\textwidth]{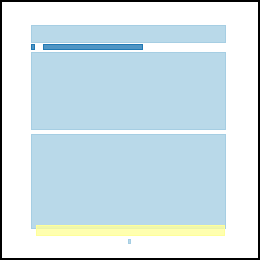} &
    \includegraphics[width=0.15\textwidth]{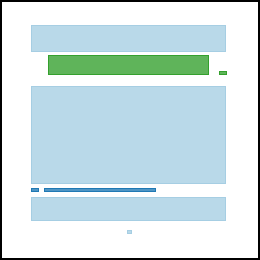} &\\     

\end{tabular}%

    \caption{\textbf{Qualitative Results for DocBank.} Competitors are poor in alignment, repeating similar patterns, containing abnormity of bounding box categories. Our method outperforms other competitors}
  \label{fig:docbank}
\end{figure*}

\begin{figure*}[t]
\centering
\setlength{\tabcolsep}{3pt}
\tiny
\begin{tabular}{p{0.08mm}cccccc}
    \rotatebox[origin=l]{90}{\textbf{LayoutVAE~\cite{jyothi2019layoutvae}}}  &
    \includegraphics[width=0.15\textwidth]{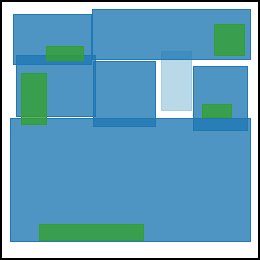} &
    \includegraphics[width=0.15\textwidth]{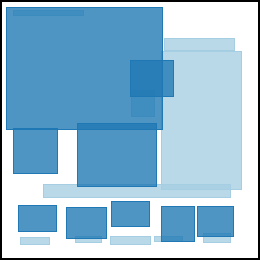} &
    \includegraphics[width=0.15\textwidth]{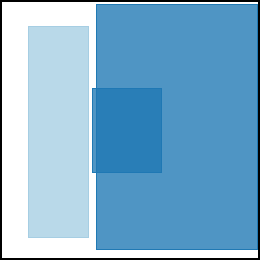} &
    \includegraphics[width=0.15\textwidth]{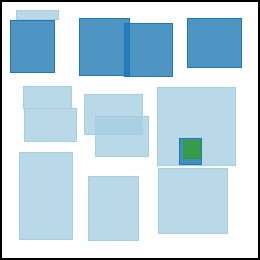} &
    \includegraphics[width=0.15\textwidth]{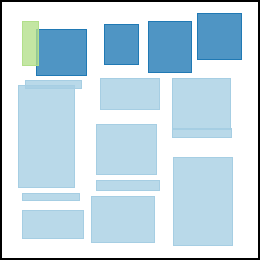} &\\

    \rotatebox[origin=l]{90}{\textbf{Gupta~\etal~\cite{gupta2021layouttransformer}}}  &
    \includegraphics[width=0.15\textwidth]{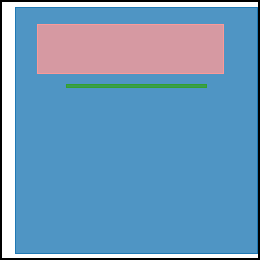} &
    \includegraphics[width=0.15\textwidth]{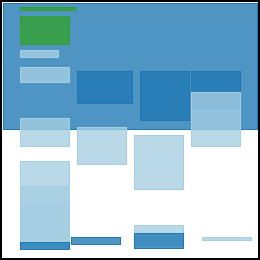} &
    \includegraphics[width=0.15\textwidth]{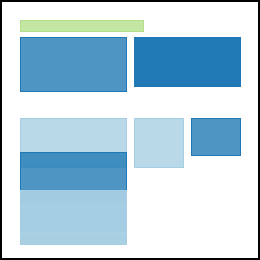} &
    \includegraphics[width=0.15\textwidth]{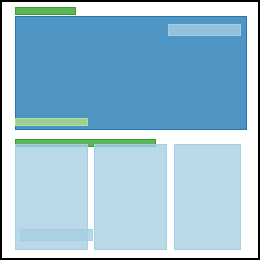} &
    \includegraphics[width=0.15\textwidth]{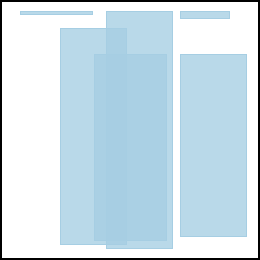} & \\

    \rotatebox[origin=l]{90}{\textbf{VTN~\cite{arroyo2021variational}}}  &
    \includegraphics[width=0.15\textwidth]{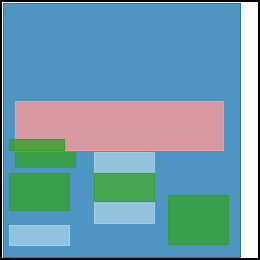} &
    \includegraphics[width=0.15\textwidth]{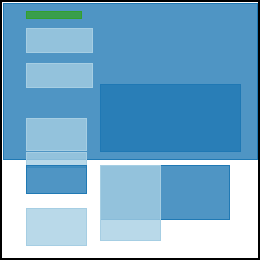} &
    \includegraphics[width=0.15\textwidth]{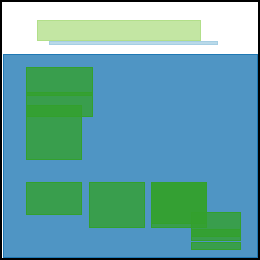} &
    \includegraphics[width=0.15\textwidth]{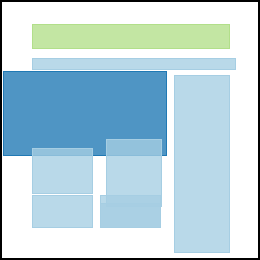} &
    \includegraphics[width=0.15\textwidth]{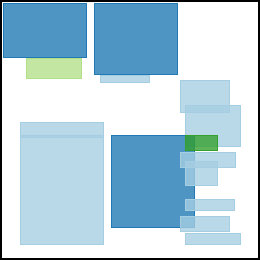} & \\

    \rotatebox[origin=l]{90}{\textbf{Ours}} &
    \includegraphics[width=0.15\textwidth]{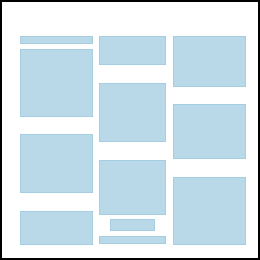} &
    \includegraphics[width=0.15\textwidth]{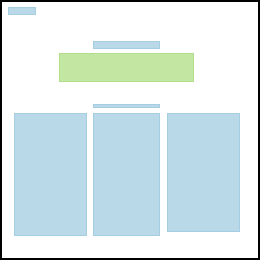} &
    \includegraphics[width=0.15\textwidth]{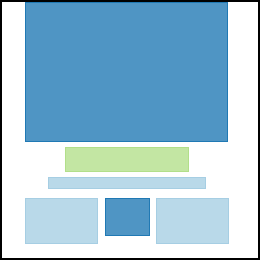} &
    \includegraphics[width=0.15\textwidth]{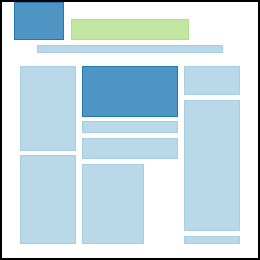} &
    \includegraphics[width=0.15\textwidth]{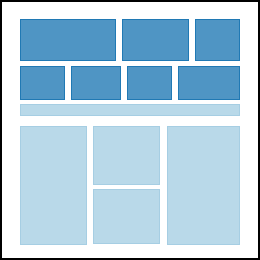} &
    \includegraphics[width=0.14\textwidth]{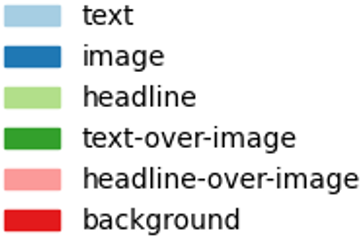} \\
\end{tabular}%

    \caption{\textbf{Qualitative Results for Magazine.} Competitors are poor in alignment, or with unreasonable overlaps. Our method outperforms other competitors.}
  \label{fig:magazine}
\end{figure*}

Figure~\ref{fig:publaynet},~\ref{fig:docbank}, and~\ref{fig:magazine} show qualitative comparison results in PubLayNet, DocBank, and Magazine dataset, respectively. The visual quality indicates that our method is able to produce diverse and realistic document layouts across three datasets. 

For PubLayNet (Figure~\ref{fig:publaynet}), LayoutVAE~\cite{jyothi2019layoutvae} generates layouts that is poor in alignment, and containing noticeable overlaps. There are significant amount of "list" categorized bounding boxes appearing in most of the generated layouts (Figure~\ref{fig:publaynet}, the second to the rightmost column in LayoutVAE row), which is not realistic to open access academic papers that PubLayNet sampled from. The similar unrealistic pattern also happens in Gupta~\etal~\cite{gupta2021layouttransformer} (Figure~\ref{fig:publaynet}, the second column in Gupta row) and VTN~\cite{arroyo2021variational} (Figure~\ref{fig:publaynet}, the rightmost column in VTN row). This abnormity will not only reduce the performance of similarity evaluation in Section~\ref{sec: quantitative}, but also negatively impact the downstream tasks using the generated layouts such as the detection task we discussed in Section~\ref{sec: downstream task}. For Gupta~\etal~\cite{gupta2021layouttransformer} and VTN~\cite{arroyo2021variational}, their performance are similar quantitatively and qualitatively. We find that if we generate layouts from the initial token rather than inputting the first bounding box (default setting by original codes), the diversity and realism of the generated results are rather repeating the same pattern (Figure~\ref{fig:publaynet}, the leftmost, the second, and the forth column in VTN~\cite{arroyo2021variational} row), or poor in alignment and overlap (Figure~\ref{fig:publaynet}, Gupta~\etal~\cite{gupta2021layouttransformer} row). However, our method outperforms other methods and generates both realistic and well-aligned document layouts without category abnormity. Our method illustrates plausible capability in both single and double column document layouts. The quality of our results will also support better performance in downstream detection task.

DocBank is the largest and the most complex dataset (more categories, diverse-sized bounding boxes) in our comparison. In this case, LayoutVAE~\cite{jyothi2019layoutvae} is unable to provide reasonable document layouts. For Gupta~\etal~\cite{gupta2021layouttransformer} and VTN~\cite{arroyo2021variational}, most of generated results are unreal and repeating patterns of permuted "equation" and "paragraph" categories (Figure~\ref{fig:docbank}, the leftmost, middle, and right most column in Gupta~\etal~\cite{gupta2021layouttransformer} row; the middle column in VTN~\cite{arroyo2021variational} row). There is also improper permutations of "reference" and 'paragraph' bounding boxes (Figure~\ref{fig:docbank}, the second column in VTN~\cite{arroyo2021variational} row). Our method shows reasonable patterns of permuted "equation" and "paragraph" categories (Figure~\ref{fig:docbank}, the leftmost, and the rightmost column in Ours row). Moreover, our method is able to handle complex design of the cover page (Figure~\ref{fig:docbank}, the second column in Ours row). And all generated layouts achieve plausible alignment.

Magazine dataset is the smallest dataset. But it has the most diverse alignment among bounding boxes (1 to 4 text columns within a single page), because the dataset is based on magazine design rather than academic papers. In this case, our competitors are unable to handle complex alignment and result in severe unreasonable overlaps between bounding boxes (Figure~\ref{fig:magazine}, the first three rows). Note that there are natural overlaps in magazine dataset since two of the categories are defined to be on top of the image category. But most overlaps are between the same category (Figure~\ref{fig:magazine}, the middle column in VTN~\cite{arroyo2021variational} row, "text-over-image" category, etc.), or unreasonable overlaps (Figure~\ref{fig:magazine}, the second column in VTN~\cite{arroyo2021variational} row, several "text" rather than "text-over-image" bounding boxes are on top of the "image" bounding boxes, etc.). However, our method is able to handle complex text alignment (Figure~\ref{fig:magazine}, the forth and the rightmost column in Ours row), and also prevent unreasonable overlaps. Our method also outperforms quantitatively in Table~\ref{tabmagazine}.

\subsection{Layout Detection Task} \label{sec: downstream task}
The best way to illustrate benefits of a generative networks is to utilize its results for downstream tasks, especially for data augmentation. Document layout detection task is a subdomain of Optical Character Recognition (OCR). The detection network is trained to segment and label each layout bounding box in the given input document images. Practically, the annotation of document layouts is tedious and time-consuming, and the accuracy of the ground truth annotation contains inevitable ambiguity. To solve these problems, we can augment the dataset by generating document images with our generated document layout designs and labels. In this way, the quality of the annotation is ensured. And we will have unlimited amount of augmented dataset for better subtask model training and evaluation.

\begin{figure*}[t]
  \includegraphics[width=\linewidth]{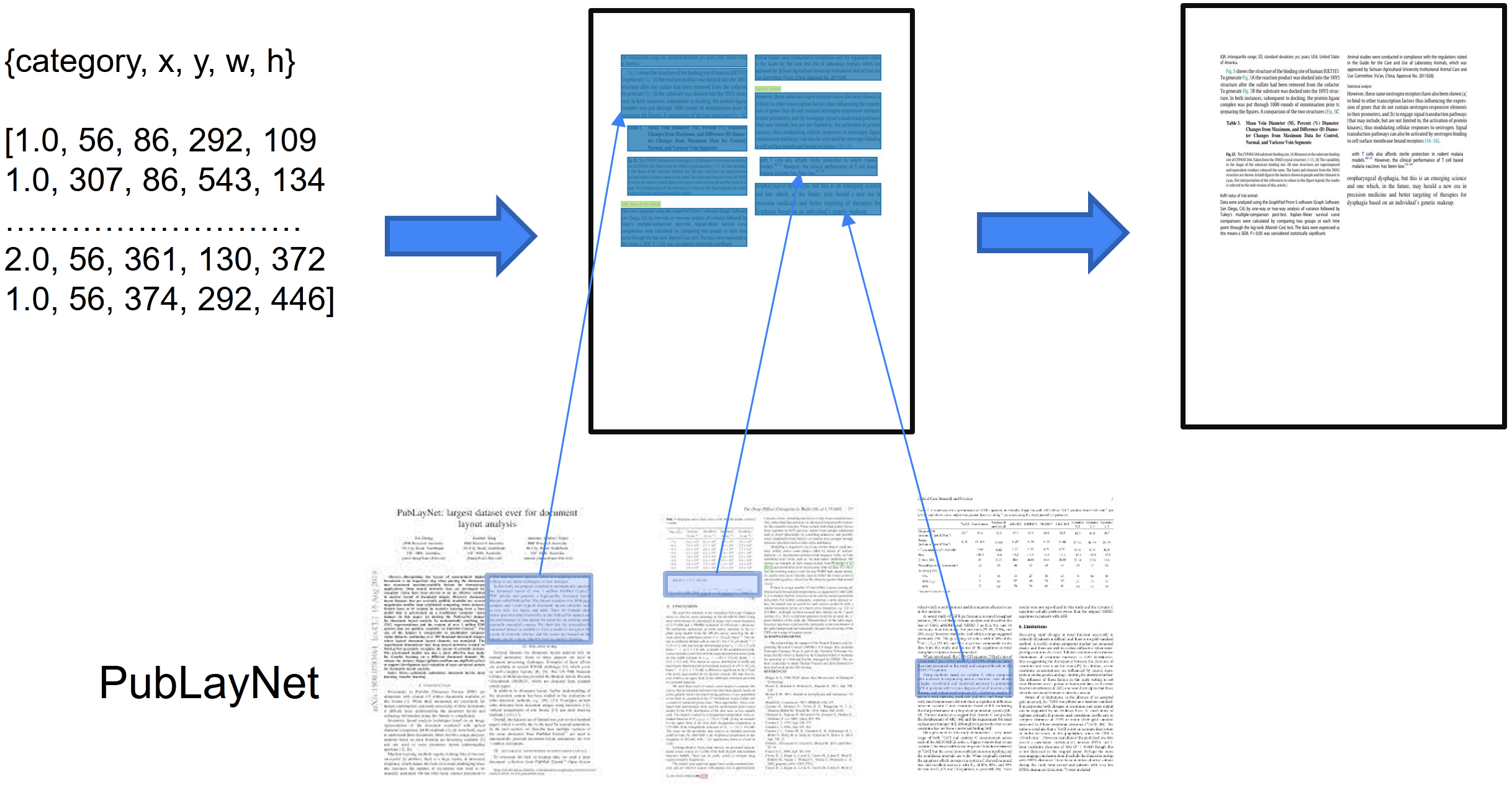}
  \caption{\textbf{Synthetic generation of document images.} Utilize our method to create a training dataset based on PubLayNet images for a layout detector. We use our generated layouts, find the most similar bounding box in real dataset (by matching category, aspect ratio, size, etc.). Then crop pixels in the original images and render a new image dataset.} 
   \label{fig:detection}
\end{figure*}

Similar to~\cite{arroyo2021variational}, as shown in Figure~\ref{fig:detection}, we develop a synthetic document image generation framework. First, we utilize our pretrained diffusion networks to randomly generate the same amount of document layouts as original training dataset (PubLayNet). Second, for each bounding box we generate, we find the nearest matching bounding box in the training dataset by matching categories, aspect ratio, size, etc. Then we slice the corresponding group of pixels from the original images in PubLayNet dataset, to synthetically mosaic a document image. Finally, we train a faster R-CNN model~\cite{ren2015faster} as our document layout detector. We compare the detection performance with the one trained by our competitor results and the original dataset for evaluation.

In Table~\ref{tabdetection}, we show the mean average precision (mAP) at IoU = 0.5. The values for VTN~\cite{arroyo2021variational} and PubLayNet are reported by its original paper. Our diffusion-based generation method enable plausible detection accuracy and outperforms VTN.

\begin{table}[!htbp] 
\begin{center}
\caption{Detection accuracy comparison between the detector trained by synthetic generated layouts, and by original PubLayNet.}\label{tabdetection}
\begin{tabular}{|l|c|c|c|}
\hline
& Ours & VTN~\cite{arroyo2021variational} & PubLayNet\\
\hline
mAP (IoU=0.5) & 0.795&  0.769 & 0.9646\\
\hline
\end{tabular}
\end{center}
\end{table}

\begin{table}[h]
\begin{center}
\caption{Ablations on Magazine Dataset}\label{ablations}
\begin{tabular}{|l|c|c|c|c|}
\hline
Ablations &  \textit{DocSim}$\uparrow$ & \textit{Doc-EMD}$\downarrow$ & Overlap & Coverage\\
\hline
 lr = 0.0001, steps = 500 & 0.156 & 0.315 & 5.34\% & 79.80\% \\
 lr = 0.0001, steps = 1000 & 0.203 & 0.245 & 3.67\% & \textbf{75.42\%} \\
 lr = 0.0002, steps = 2000 & 0.282 & 0.172 & 1.56\% & 72.34\% \\
 lr = 0.00001, steps = 2000 & 0.274 & 0.133 & 2.33\% & 71.21\% \\
 Ours (lr = 0.0001, steps = 2000) & \textbf{0.302} & \textbf{0.117} &  \textbf{1.23\%} & 70.55\% \\
\hline
 Real Data & & & 1.36\% & 76.00\% \\
\hline
\end{tabular}
\end{center}
\end{table}

\subsection{Ablations}
In our model training, we conduct an ablation study on both learning rate and diffusion steps, as shown in Table~\ref{ablations}. In general, more diffusion steps will improve network performance. But more diffusion steps also lead to extremely longer training time. We found that more diffusion steps above 2000 have no explicit benefits to the performance. Thus we choose diffusion steps as 2000 for all our trainings on Magazine and other datasets. Meanwhile, a proper learning rate is also the key to good performance. Though we found that diffusion model performance is quite robust to different learning rate. An arbitrary learning rate may still negatively influence the network performance.

\section{Conclusion}
In this work, we develop a new approach for document layout generation and a novel metric for document layouts evaluation. Our diffusion-based document layout generation approach shows outperforming and competitive results on several well known datasets in document layout generation. We have also shown how our approach can be applied to downstream applications, such as pretraining a document layout detector. In this paper, we provide extensive qualitative analysis and examples, so our readers understand both limitations and advantages of our method and our proposed metric. 

In the future work, we plan to explore domain generalization and conditional generation. For domain generalization, higher specificity layouts as well as full OCR generation at the line level would provide a more complete document generation system. Meanwhile, conditioning generation on the right set of factors allows the proposed method to be used more effectively in document generation pipelines.



%
%
%
\bibliographystyle{splncs04}
\bibliography{mybibliography}

\end{document}